\documentclass[letterpaper]{article} 
\usepackage{aaai2026}  
\usepackage{times}  
\usepackage{helvet}  
\usepackage{courier}  
\usepackage[hyphens]{url}  
\usepackage{graphicx} 
\usepackage{natbib}  
\usepackage{caption} 
\usepackage{algorithm}
\usepackage{algorithmic}

\usepackage{newfloat}
\usepackage{listings}
\usepackage{algorithm}
\usepackage{algorithmic}
\usepackage{xcolor}
\usepackage{multirow}
\usepackage{booktabs}
\usepackage{amsmath}

\usepackage{afterpage, makecell, booktabs, multirow}
\usepackage{float}
\usepackage{tabularray} 
\UseTblrLibrary{booktabs} 
\usepackage{afterpage} 
\DeclareCaptionStyle{ruled}{labelfont=normalfont,labelsep=colon,strut=off} 
\floatstyle{ruled}
\newfloat{listing}{tb}{lst}{}
\floatname{listing}{Listing}
\title{Collaborative Medical Triage under Uncertainty: A Multi-Agent Dynamic Matching Approach}
\author{
    Hongyan Cheng,
    Chengzhang Yu,
    Yanshu Shi,
    Chiyue Wang,
    Cong Liu,
    Zhanpeng Jin,
}
\affiliations{
    \textsuperscript{\rm 1}South China University of Technology\\

}

\usepackage{bibentry}

\begin{document}

\maketitle

\begin{abstract}
The post-pandemic surge in healthcare demand, coupled with critical nursing shortages, has placed unprecedented pressure on medical triage systems, necessitating innovative AI-driven solutions. We present a multi-agent interactive intelligent system for medical triage that addresses three fundamental challenges in current AI-based triage systems: inadequate medical specialization leading to misclassification, heterogeneous department structures across healthcare institutions, and inefficient detail-oriented questioning that impedes rapid triage decisions. Our system employs three specialized agents—RecipientAgent, InquirerAgent, and DepartmentAgent—that collaborate through Inquiry Guidance mechanism and Classification Guidance Mechanism to transform unstructured patient symptoms into accurate department recommendations. To ensure robust evaluation, we constructed a comprehensive Chinese medical triage dataset from ``Ai Ai Yi Medical Network'', comprising 3,360 real-world cases spanning 9 primary departments and 62 secondary departments. Experimental results demonstrate that our multi-agent system achieves $89.6\%$ accuracy in primary department classification and $74.3\%$ accuracy in secondary department classification after four rounds of patient interaction. The system's dynamic matching based guidance mechasnisms enable efficient adaptation to diverse hospital configurations while maintaining high triage accuracy. Our work provides successfully developed this multi-agent triage system that not only adapts to organizational heterogeneity across healthcare institutions but also ensures clinically sound decision-making.
\end{abstract}


\section{Introduction}
Currently, global healthcare systems are facing dual pressures. On one hand, emerging and re-emerging infectious diseases such as influenza and dengue fever continue to impact public health systems, creating significant challenges for disease prevention and control mechanisms \cite{sharma2025future}. On the other hand, the rapid rise in non-communicable diseases (e.g., cardiovascular diseases and diabetes) due to population aging and lifestyle changes has further exacerbated medical resource shortages and economic burdens through long-term treatment demands \cite{awofala2024data}.Consequently, healthcare systems are experiencing unprecedented operational pressures post-pandemic. For instance, the NHS experienced an 8.1\% increase in outpatient appointments in 2023–24, significantly straining hospital capacity and resources \cite{NHS2024report}. Despite investments to expand services, rising outpatient demand continues to challenge timely care delivery. This imbalance between rising patient demand and an insufficient nursing workforce places immense strain on conventional triage systems. The imbalance between surging patient demand and an insufficient nursing workforce exerts tremendous pressure on conventional triage systems. A study conducted at a tertiary medical center with an annual caseload of approximately 70,000 patients confirmed that delays in both door-to-triage and triage-to-patient administration processes are key exacerbating factors\cite{KIENBACHER2022219}. 

Against the backdrop of millions of patients awaiting consultations amidst a severe shortage of clinical resources, ensuring each patient-provider interaction is both precise and efficient has become critically important. Modern healthcare is an increasingly complex and fragmented system, in which patients often face challenges connecting with the appropriate specialty at the initial stage of seeking help \cite{budde2021role}. Any incorrect initial referral not only wastes valuable clinical resources and delays effective treatment for the patient, but also further exacerbates the operational burden on the entire system. Therefore, optimizing the departmental routing process—by accurately interpreting patient complaints through intelligent means and guiding them to the correct department—holds critical and urgent practical significance for alleviating system bottlenecks and enhancing overall healthcare service efficiency.

In fact, large language models like GPT-3 have demonstrated significant potential in generating accurate preliminary diagnoses and triage recommendations by analyzing patients' symptom descriptions. Furthermore, AI-based triage systems have been implemented in outpatient clinics, enabling optimized diagnostic pathways through intelligent analysis of patient data and dynamic allocation of medical resources, thereby improving the accuracy and consistency of triage.

However, existing agent-based medical triage systems encounter three significant challenges that limit their effectiveness in clinical practice. The first challenge is inadequate medical specialization. This leads to general intelligent agents, without specialized medical fine-tuning, generating erroneous triage recommendations that could potentially cause unnecessary harm to patients and delay critical treatment processes. The second challenge involves the heterogeneous department structures across different healthcare institutions, where large hospitals maintain highly specialized departments such as pancreatic surgery and vascular surgery, while smaller hospitals lack such granular specialization due to resource constraints. Although common approaches include fine-tuning large language models or implementing Retrieval-Augmented Generation (RAG) technologies to adapt to specific hospital configurations, model fine-tuning incurs prohibitive costs for most healthcare institutions, while native RAG systems with coarse-grained data repositories fail to effectively accommodate the diverse organizational structures of different hospitals. The third challenge concerns the efficiency requirements of medical triage processes, which demand rapid decision-making to minimize patient wait times and expedite appropriate department assignment, necessitating agents with macro-level reasoning capabilities. However, current large language models exhibit a tendency toward excessive detail-oriented questioning that impedes efficient triage flow, such as engaging in prolonged inquiries about specific fever temperatures or symptom duration when patients report fever symptoms, despite these granular details contributing minimally to accurate department classification.

We developed a multi-agent interactive intelligent triage system to address these challenges. The core innovation lies in the synergistic interaction of multiple agents through meticulously engineered prompts to enhance specialized medical capabilities. The architecture is further optimized through dynamic inquiry guidance and classification guidance mechanisms. To ensure adaptability to varying departmental structures across different hospitals, we constructed a large-scale dataset sourced from the ``Ai Ai Yi Medical Network''\footnote{\url{https://www.iiyi.com/}} (as opposed to single-institution data). This dataset encompasses 9 primary departments and 62 secondary departments, enabling the system to accommodate diverse hospital requirements. The multi-source data design effectively validates the system's generalization capability across different clinical scenarios and organizational structures.

\section{Related Work}

\subsection{Triage Datasets}
Most existing triage datasets are derived from electronic medical records (EMRs) of single institutions, while the inherent heterogeneity in the structure and annotation systems across different departments constitutes a key barrier to the generalization of artificial intelligence models. Classification standards for the same symptoms often vary between hospitals \cite{Cui2019SymptomatologyDO} (for example, "headache" may be categorized under "Internal Medicine" or "Neurology") \cite{Kolosova1996TheCO}. Moreover, differences in department setups, diagnostic and treatment workflows, and patient population distributions lead to highly heterogeneous data characteristics\cite{Henkel2019StructuredDA, Evans2021ARO}. Such inconsistencies in labeling and structure severely undermine the model’s ability to generalize across institutions, making it difficult for models to maintain stable predictive performance in new medical environments \cite{Behar2023GeneralizationIM}. At the same time, noise, missing data, and variations in coding standards further exacerbate the complexity of model training and increase the risk of bias \cite{hofer2021ai, huang2021gloria}.

\subsection{Multi-Agent-Based Triage Systems}
Artificial intelligence technology has driven the evolution of medical triage systems. Early rule-based expert systems relied on fixed knowledge bases and decision trees; although effective for structured symptoms, they struggled to handle the uncertainty and ambiguity present in clinical consultations \cite{rulebased2018}. With the development of machine learning, more systems began leveraging real-world data such as electronic health records (EHR) to predict patient triage levels and care pathways through supervised learning \cite{wang2024recent}. However, these “passive” systems depend on standardized inputs, lack real-time interaction with patients, and have difficulty adapting to complex contextual changes \cite{SHEKHAR202527}.

In recent years, agent-based models have shifted triage systems from static prediction to active interaction. Conversational agents simulate clinical interviews through multi-turn natural language dialogues—for example, systems leveraging advanced retrieval-augmented large language models have demonstrated improved knowledge integration and decision accuracy in clinical settings \cite{wang2024blendfilter}. With the advancement of LLMs such as ChatGPT and Med-PaLM, triage agents have significantly improved in language understanding and reasoning, enabling more flexible handling of open-ended questions and differential diagnosis \cite{han2024}. Building on this, multi-agent systems have been developed to further optimize triage by collaborative multi-tasking. The TriageAgent framework proposed by \citet{lu2024triageagent}. employs multiple LLM-based agents working together, greatly enhancing system robustness and scalability.

\section{Method}

\subsection{Overall Architecture}
Our triage system is achieved through three tightly integrated core components. First, the data processing module employs LLMs to impute missing information, delivering standardized inputs for subsequent analysis. Next, a multi-agent system—comprising RecipientAgent, InquirerAgent, and DepartmentAgent—collaborates under Inquiry Guidance and classification guidance mechanisms to iteratively refine illness history collection and optimize department recommendations through dynamic iterations. Finally, Evaluate system was used to quantify the entire workflow. A complete system was built, encompassing data processing, intelligent interaction, and results evaluation.

\subsection{Data Processing}

To construct a high-quality dataset for the intelligent triage model, we initially collected 8,769 real electronic health records (EHR) from  ``Ai Ai Yi Medical Network''\footnote{\url{https://www.iiyi.com/}}.

To ensure data quality, we first implemented a rigorous data cleaning pipeline: removing completely duplicate records and filtering out templated invalid data with highly consistent content in key fields such as chief complaints and History of Present Illness. Building upon this, to address the persistent issue of missing key fields in the cleaned data, we employed a large language model (Qwen-plus) based approach. Through three key steps , Integrity detection, Prompt Construction, and LLM-based completion . we systematically imputed the data, effectively ensuring record completeness.

Ultimately, we constructed a final dataset containing 3,360 high-quality EHRs. This dataset not only includes key information such as patient demographics, chief complaints, and present illness history with their actual department assignments. The dataset spans 9 primary departments (e.g., Internal Medicine, Surgery, Pediatrics, Obstetrics \& Gynecology) and 62 secondary departments (e.g., Neurology, Nephrology, Endocrinology, and Respiratory Medicine under Internal Medicine). providing a comprehensive foundation for intelligent triage system development.

\subsection{Multi-agent Systems}
\begin{figure*}[!htb]
\centering
\includegraphics[height=0.4\textheight
]{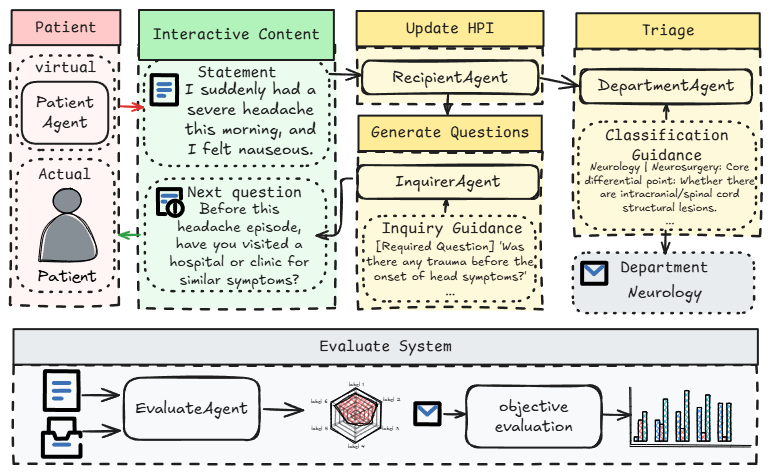}
\caption{Architecture of the Multi-Agent Collaborative Triage System. The figure illustrates the dynamic workflow where the Patient module (red) and the core agent system (yellow) collaborate through multi-round interaction (green). The core agents—RecipientAgent, InquirerAgent, and DepartmentAgent—are responsible for symptom processing, question generation, and department recommendation, with their operational logic defined by Equations (\ref{eq:recipient_agent})–(\ref{eq:inquier}). The evaluation module (gray) performs a dual evaluation based on six-dimensional metrics and the accuracy of the department recommendation.}
\label{fig:figure1}
\end{figure*}

We propose a multi-agent collaborative triage architecture in Figure \ref{fig:figure1} , the core of which consists of three agents: The RecipientAgent transforms patients' unstructured symptom information into standardized History of Present Illness (HPI) records, the InquirerAgent identifies missing critical information in the HPI and conducts targeted questioning of patients, and the DepartmentAgent performs precise department recommendation decisions based on complete HPI data.

The system operates with support from large language models(LLM), where agent invocation timing is determined by iterative interaction rounds. When the system fails to reach the preset round limit, it automatically triggers multiple inquiry cycles. The InquirerAgent and DepartmentAgent form a dynamic feedback loop that continuously optimizes information collection and department differentiation accuracy, thereby enhancing the final department recommendation precision.

\subsection{RecipientAgent}

The core responsibility of the RecipientAgent is to systematically transform a patient's unstructured, colloquial descriptions into a standardized History of Present Illness (HPI). In each interaction round $t$, it integrates three key inputs: the patient's current description ($D_t$), the question from the previous round ($q_{t-1}$), and the accumulated historical HPI ($H_{t-1}$).

This process not only handles explicit information directly stated by the patient but also extracts implicit clues from the physician's questions, ensuring coherence and completeness by integrating with historical records. Ultimately, it generates a complete and structured HPI for the current round, $H_t$ . This process is formally represented by Equation~(\ref{eq:recipient_agent}):
\begin{equation}
    H_t = R(D_t, q_{t-1}, H_{t-1})
    \label{eq:recipient_agent}
\end{equation}

where $R(\cdot)$ represents the RecipientAgent. The generated $H_t$ provides a robust data foundation for the subsequent stages of diagnostic reasoning and triage.

\subsection{DepartmentAgent}

The DepartmentAgent serves as the decision-making core of the triage system. Its primary task in each interaction round $t$ is to generate both an optimal department recommendation, $d_t$, and a set of candidate departments for differentiation, $\hat{d}_t$. This dual output is based on the standardized History of Present Illness (HPI), $H_t$; the list of available departments, $C$; and any dynamic departmental guidance, $G_k$. The agent operates on a "Macro-Judgment Principle" . When Classification Guidance $G_k$ is available, the agent prioritizes key directives such as "exclude Department XX" or "vs. Department XX" to resolve ambiguity between similar departments. This dual-output process can be represented by a unified decision function, $A(\cdot)$:
\begin{equation}
    (d_t, \hat{d}_t) = A(H_t, C, G_k)
    \label{eq:dept_agent_unified}
\end{equation}
where The $\hat{d}_t$, generated only when high ambiguity exists in the department selection. Its purpose is to provide a clear signal to the InquirerAgent for the next round of questioning. Through this dual-output mechanism, the system not only provides a clear judgment ($d_t$) in each round but also proactively plans the next information-gathering strategy via $\hat{d}_t$, forming an efficient, dynamic optimization loop. The final triage recommendation is the $d_t$ produced in the final round of interaction.

\subsection{InquirerAgent}

The core task of the InquirerAgent is to analyze the existing History of Present Illness (HPI), $H_t$, and the sequence of historical questions, $\mathcal{Q}_{t-1}$, in order to dynamically generate the next question, $q_t$. Its questioning strategy strictly adheres to a principle of no repetition and high specificity, adapting its approach based on the consultation's context.For example, if the HPI mentions patient headache but does not specify headache frequency and triggering factors, the InquirerAgent focuses on these missing parts, generating targeted questions such as "Do you experience blurred vision or tinnitus during headaches?" This process can be represented by Equation~(\ref{eq:3}), (\ref{eq:inquier}):
\begin{equation}
    q_1 = I(H_1, \hat{d}_1)
    \label{eq:3}
\end{equation}
\begin{equation}
    q_t = I(H_t, \mathcal{Q}_{t-1}, \hat{d}_t)
    \label{eq:inquier}
\end{equation}
where $I(\cdot)$ represents the InquirerAgent. The input $\mathcal{Q}_{t-1} = \{q_1, q_2, \dots, q_{t-1}\}$ denotes the entire sequence of historical questions. The function utilizes the candidate departments $\hat{d}_t$ as an optional guidance signal to ensuring both the completeness of medical history information and effective differentiation between target departments.

when guided by a list of candidate departments, $\hat{d}_t$, the agent pivots to generating questions aimed at differentiation. To illustrate, to distinguish between Neurology and Neurosurgery, it might ask a highly discriminative question like, "Is the headache accompanied by projectile vomiting?"

\subsection{Inquiry Guidance Mechanism}

The Inquiry Guidance mechanism operates through a series of department-specific Knowledge Bases (KBs), where each medical specialty maintains its own curated set of rules and heuristics. Each KB is structured around core decision-making components, including Core Inquiry to identify key differential questions (e.g., for Internal Medicine, determining if a symptom requires medication over surgical treatment); Avoid Detail Entanglement to maintain focus on triage rather than diagnosis by avoiding overly specific questions (like pain severity); Exclusion Rules to define specialty exclusion criteria (e.g., ruling out Surgery by inquiring about trauma history); and Secondary Department Differentiation to establish sub-specialty selection logic (e.g., distinguishing Neurology from Neurosurgery based on trauma history).

This mechanism is primarily applied in the InquirerAgent to ensure precision in the line of questioning through these department-specific rules. For instance, when a patient presents with chest pain, Internal Medicine rules prioritize asking about a history of hypertension or diabetes to assess cardiovascular needs. Conversely, if a trauma history is mentioned, Surgery rules immediately trigger emergency screening questions, such as inquiring about breathing difficulties, thereby reducing ineffective inquiries and targeting high-risk triage pathways.

\subsection{Classification Guidance Mechanism}
The Classification Guidance mechanism uses a rule engine built upon department-specific Knowledge Bases (KBs), combining their differentiation rules with the dynamic prompts of the DepartmentAgent to generate precise recommendations. The mechanism defines detailed departmental comparison rules (covering symptoms, surgical indications, and medication needs) and assigns priority levels to prioritize high-weight features in complex scenarios. For instance, rules distinguish between Gastroenterology and General Surgery, where "acute abdominal pain with peritoneal irritation signs" indicates General Surgery, while "chronic gastritis" is classified under Gastroenterology. Similarly, for differentiating Neurology from Neurosurgery, a patient with "sudden severe headache with vomiting but no trauma history" leads to a Neurology recommendation after Neurosurgery is excluded due to the lack of trauma.

\subsection{Evaluate System}
In the system, we measure its performance through two types of evaluations. The first is objective evaluation, where we compare the optimal department recommendation of DepartmentAgent, $d_t$, with the actual department from the case to determine its accuracy. This includes the accuracy for primary departments, secondary departments, and the overall accuracy. The second is a six-dimensional evaluation using the EvaluateAgent. The main goal of this evaluation is to assess the intelligent system's performance in clinical scenarios through multi-dimensional quantitative analysis, thereby providing data-driven support for system optimization. The six evaluation dimensions include: the system's inquiry logic, triage decision-making, diagnostic reasoning, communication capabilities, multi-round interaction consistency, and overall professional competency.with detailed scoring criteria for each dimension presented in Table~\ref{tab:evaluation_criteria}. The scoring system ranges from 1 to 5 points, where higher scores indicate superior performance across different clinical competencies.

\begin{table*}[htbp]
\centering
\small
\begin{tabular}{p{1.6cm}>{\centering\arraybackslash}p{0.8cm}p{4.7cm}p{7.9cm}}
\toprule
\textbf{Dimension} & \textbf{Score} & \textbf{Description} & \textbf{Example Criteria} \\
\midrule
\multirow[t]{3}{1.6cm}{Clinical Inquiry Capability} 
& 1 & Completely non-targeted inquiry & Random questioning without focus \\
& 3 & Basic systematic collection & Covers main symptoms systematically \\
& 5 & Comprehensive systematic collection & Follow-up on pain characteristics, duration, and MI features \\
\midrule
\multirow[t]{3}{1.6cm}{Triage Accuracy} 
& 1 & Completely incorrect & Wrong primary department selection \\
& 3 & Partially correct & Correct primary, suboptimal secondary \\
& 5 & Optimal choice & Subarachnoid hemorrhage to neurosurgery \\
\midrule
\multirow[t]{3}{1.6cm}{Diagnostic Reasoning Ability} 
& 1 & Completely erroneous reasoning & Illogical diagnostic process \\
& 3 & Basic reasoning with gaps & Partial differential diagnosis \\
& 5 & Rigorous and complete reasoning & Abdominal pain excluding GI/urological/gynecological causes \\
\midrule
\multirow[t]{3}{1.6cm}{Communication Expression} 
& 1 & Unclear expression & Confusing medical terminology \\
& 3 & Generally clear & Basic professional communication \\
& 5 & Perfect expression & Balance between professionalism and accessibility \\
\midrule
\multirow[t]{3}{1.6cm}{Multi-round Consistency} 
& 1 & Logical confusion & Contradictory information usage \\
& 3 & Generally coherent & Some information integration \\
& 5 & Completely coherent & Progressive deepening inquiry \\
\midrule
\multirow[t]{3}{1.6cm}{Overall Professionalism} 
& 1 & Very low professional level & Poor clinical judgment \\
& 3 & Adequate professional level & Standard clinical competency \\
& 5 & Expert-level competency & Recognition of rare critical conditions \\
\bottomrule
\end{tabular}
\caption{Evaluation Criteria for Multi-dimensional Clinical Assessment}
\label{tab:evaluation_criteria}
\end{table*}

\subsection{PatientAgent}

The PatientAgent is an intelligcent agent built upon real Electronic Health Records (EHR) that engages in interactive content with the InquirerAgent. The agent's behavior is grounded in an authentic Electronic Health Record (EHR). The PatientAgent adheres to a Progressive Information Disclosure principle. In the first round of dialogue, it only concisely states the core chief complaint. In subsequent rounds, it provides targeted responses strictly based on the question ($q_t$) posed by the InquirerAgent, and can also simulate human-like uncertainty. Its input-output relationship is represented by Equation~\ref{eq:4}:
\begin{equation}
    R_t = \text{PatientAgent}(q_t, E_t, P)
    \label{eq:4}
\end{equation}
where $P$ represents the structured patient EHR, $q_t$ is the next question from the InquirerAgent, $E_t$ is the dialogue round indicator, and $R_t$ is the generated patient response.

\section{Experiment}
\subsection{Experimental Setup}
We evaluate our multi-agent system on the dataset described in Section~3.2 through a simulated four-round doctor-patient interaction. The system consists of three core collaborating agents (RecipientAgent, InquirerAgent, and DepartmentAgent), with their decisions are driven by the DeepSeek-V3 large language model, with both the Inquiry Guidance and Classification Guidance mechanisms enabled. The performance is assessed by the Evaluate System through a dual evaluation: first, by evaluating triage accuracy, and second, by using an EvaluateAgent to quantify the quality of the entire process.

\subsection{Overall Performance Evaluation}

\begin{figure}[htbp]
    \centering
    \includegraphics[width=0.5\textwidth, height=5cm]{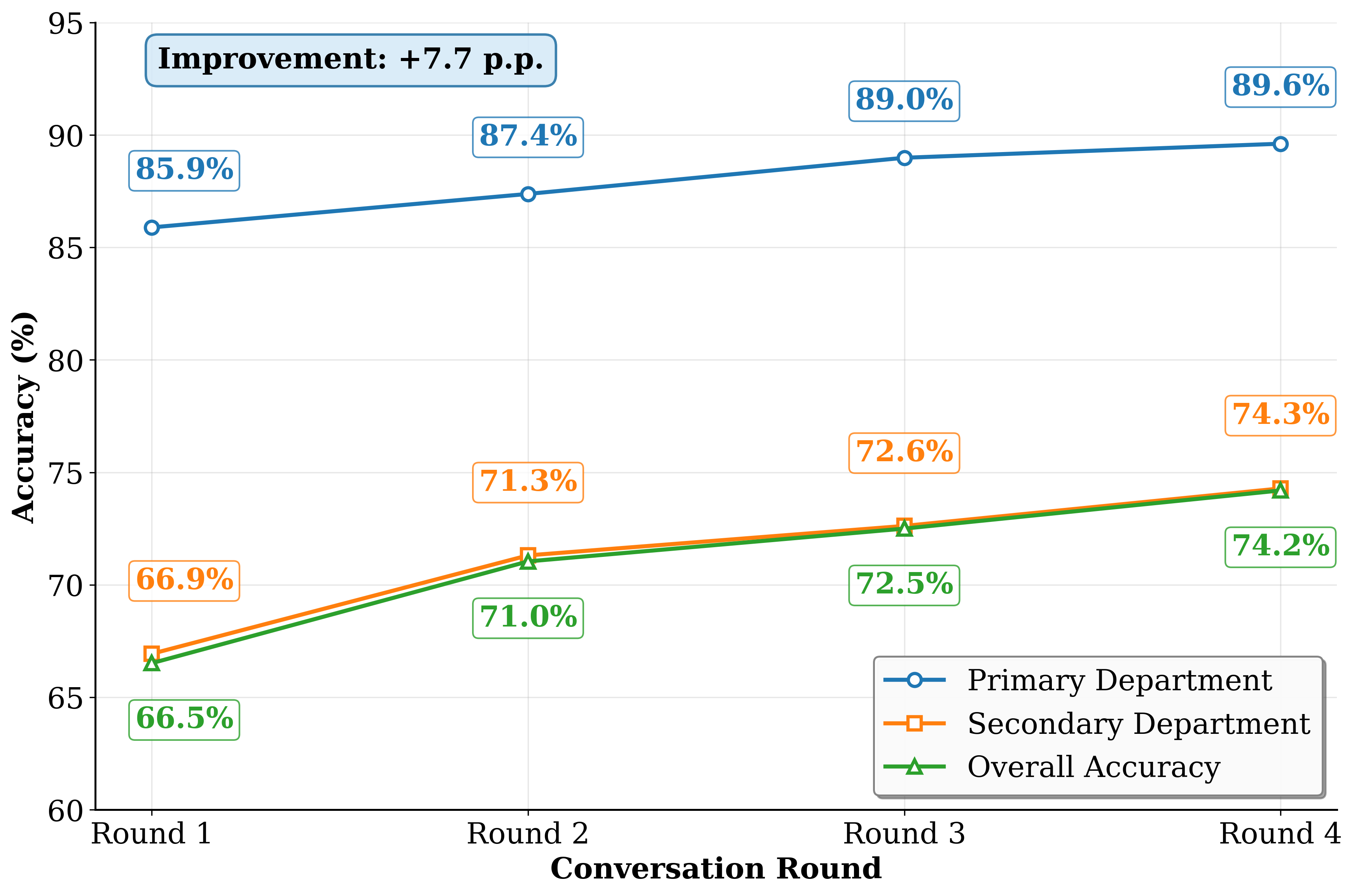} 
    \caption{Overall Accuracy rose by +7.7\%(to 74.2\%) across four rounds.}
    \label{fig:figure2}
    \vspace{0pt}
\end{figure}

We first conducted a comprehensive evaluation of the system's overall performance on 3,360 real-world clinical cases. As shown in Figure~\ref{fig:figure2}, our system demonstrates remarkable dynamic learning capabilities. Through four rounds of interaction, its Overall Accuracy achieved a net improvement of +7.7\% , rising from 66.5\% to 74.2\%. This was driven by steady gains in both Primary Department accuracy (85.9\% to 89.6\%) and Secondary Department accuracy (66.9\% to 74.3\%). These results robustly validate that our multi-round interaction mechanism can progressively refine triage decisions through active information gathering.

\subsection{Multidimensional Capability Assessment}
\begin{figure}[htbp]
    \centering
    \includegraphics[width=0.8\linewidth,height=5cm]{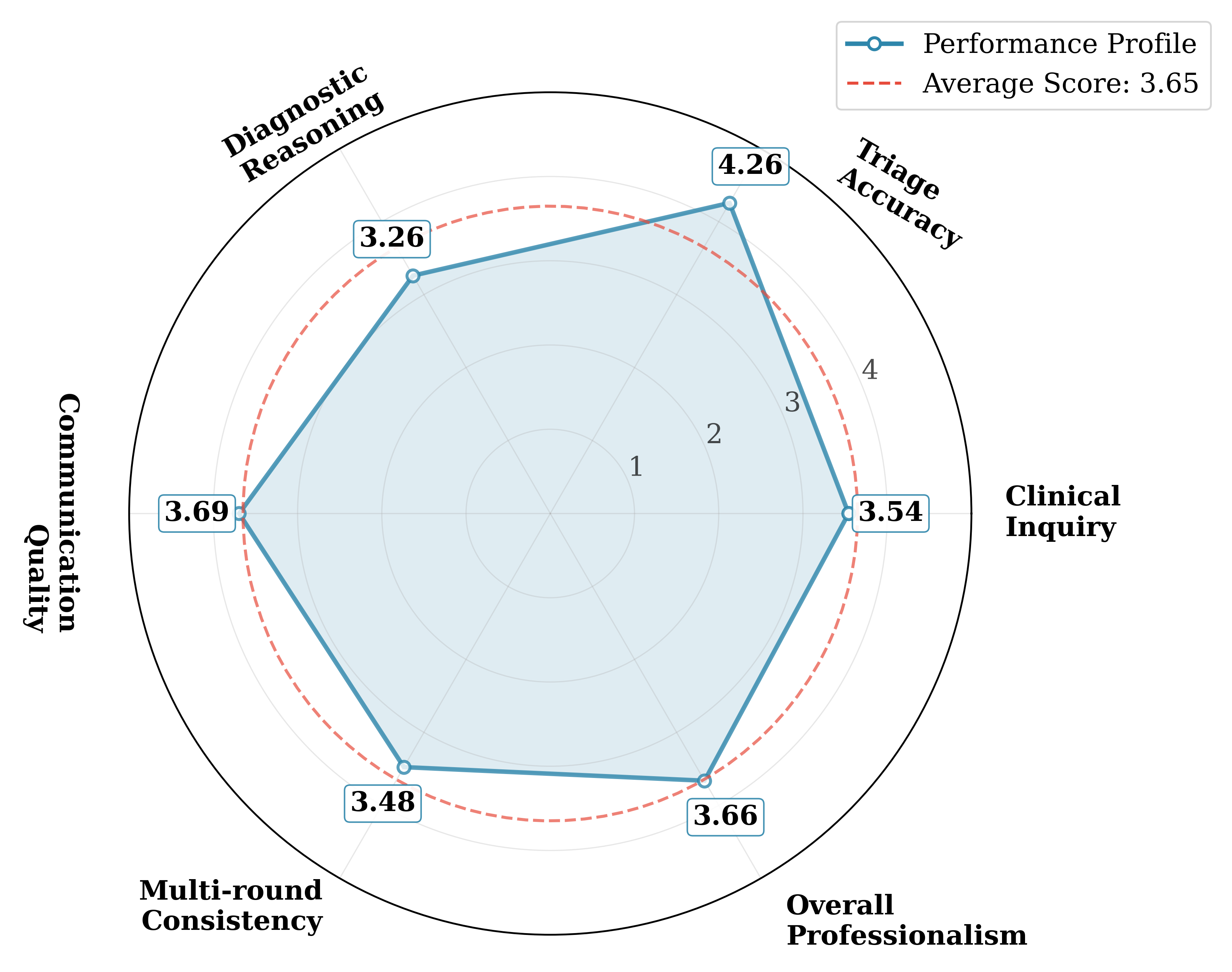}
    \caption{Multidimensional Clinical Capability Assessment. The radar chart visualizes the system's average performance across six key clinical dimensions on a 1-to-5 scale. }
    \label{fig:figure3}
    \vspace{0pt}
\end{figure}

To assess its clinical applicability, we conducted a multidimensional capability assessment of the system in Figure~\ref{fig:figure3}.
The system achieved a solid average score of 3.65 out of 5 across six dimensions. Its core capability, Triage Accuracy, scored the highest at 4.26. The system also demonstrated balanced performance in dimensions such as Communication Quality (3.69) and Overall Professionalism (3.66), with Diagnostic Reasoning (3.26) identified as the primary area for future optimization.

\subsection{System Reliability}
\begin{figure}[htbp]
    \centering
    \includegraphics[width=0.5\textwidth]{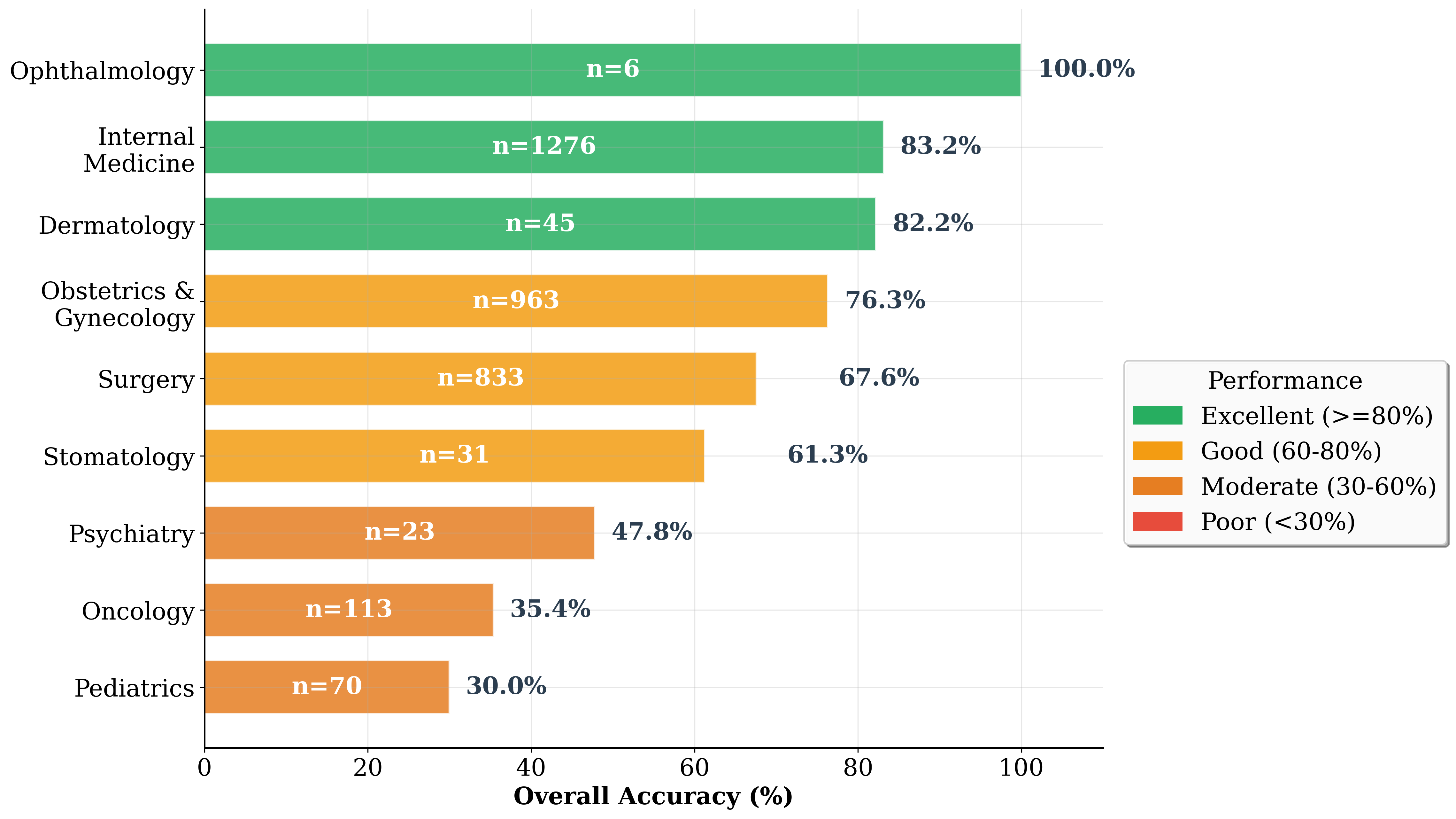}
    \caption{Department-wise Triage Accuracy. Final-round accuracy by primary department, annotated with sample size (n). The dashed line indicates the average accuracy of 64.9\%.}
    \label{fig:figure4}
    \vspace{0pt}
\end{figure}

To validate the triage reliability of our system, we conducted a department-wise analysis in Figure~\ref{fig:figure4}. The results demonstrate that with robust performance in high-sample specialties like Internal Medicine (83.2\%) and relatively weaker performance in complex, high-sample specialties such as Pediatrics (30.0\%). Notably, the system exhibits excellent fault tolerance by maintaining accurate identification of the primary specialty even when triage deviations occur.

\subsection{Error Analysis}
\begin{figure}[htbp]
    \centering
    \includegraphics[width=0.5\textwidth]{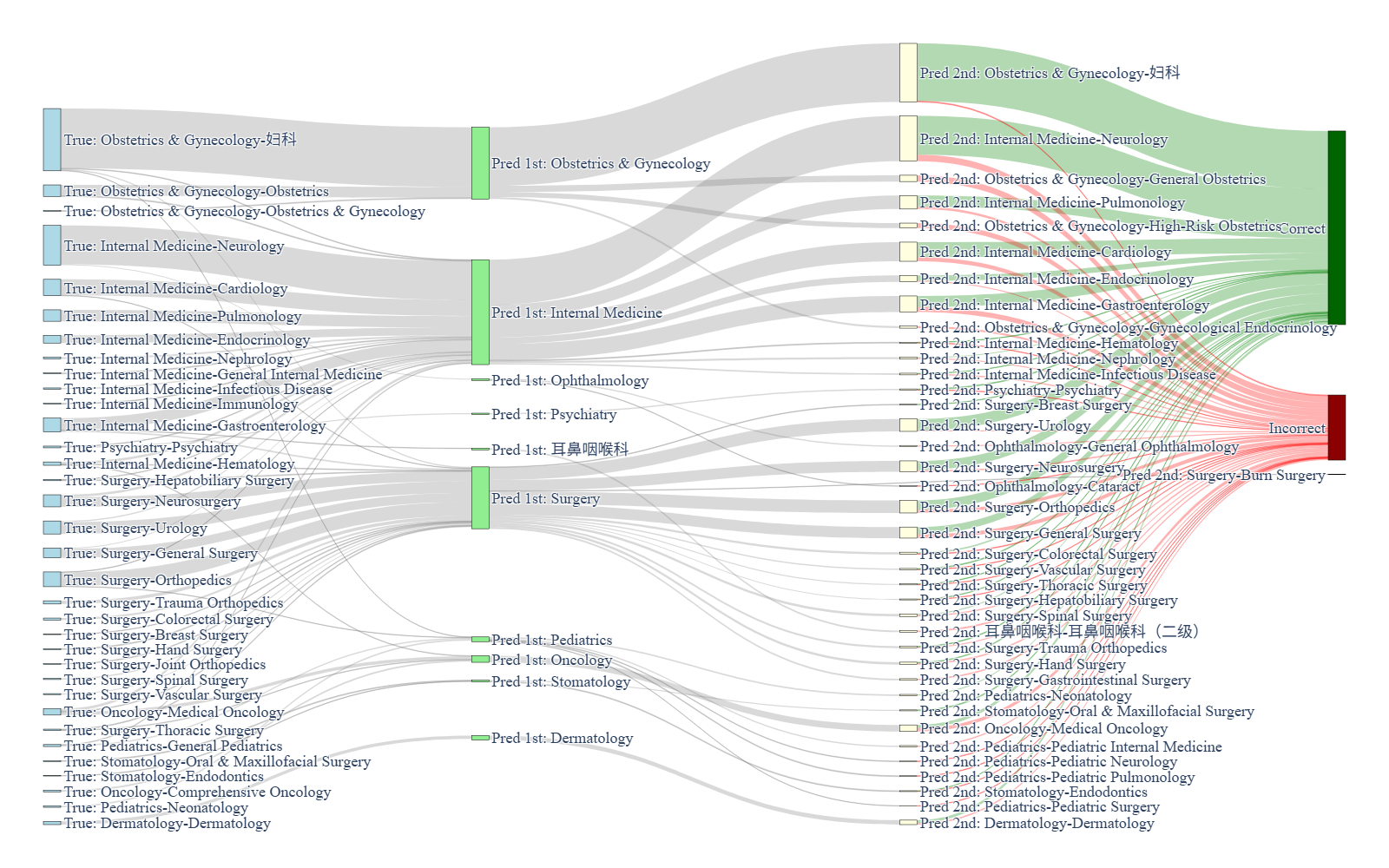} 
    \caption{ Sankey Diagram of Triage Misclassification Pathways (n=3,660). Flow width is proportional to case volume, with green and red paths indicating correct and incorrect outcomes.}
    \label{fig:figure5}
    \vspace{0pt}
\end{figure}

For an in-depth root cause analysis, we conducted a thorough error attribution analysis. we visualized misclassification pathways using a Sankey diagram in Figure~\ref{fig:figure5}. The analysis revealed significant cross-specialty misclassification flows, particularly bidirectional errors between Internal Medicine and Surgery, along with an important error pattern distinction: low-risk secondary department errors (misclassifications within correct primary departments) accounted for 15.4\% of cases, substantially exceeding the 10.4\% rate of critical primary department errors. These findings demonstrate the system's strong reliability in core specialty identification while highlighting specific opportunities for improving sub-specialty classification accuracy.

\subsection{Ablation Studies on Core Mechanism Effectiveness}
To investigate the fundamental drivers behind our system's strong performance, we constructed and defined a specialized test set of "challenging yet solvable" cases. The criteria for this test set are explicitly defined as: Cases where the system's initial prediction (in the first round) was incorrect, but which were successfully corrected by the final round of interaction.Building upon our system's overall demonstrated effectiveness, we conducted a series of in-depth ablation studies using this curated set of 258 cases.

\subsection{The Necessity of RecipientAgent}
We first aimed to validate the foundational role of the RecipientAgent within our architecture. To this end, we compared our Model against a No HPI Model, an ablated version that bypasses the RecipientAgent and uses only raw, concatenated dialogue as its input.

\begin{figure}[htbp]
    \centering
    \includegraphics[width=0.5\textwidth]{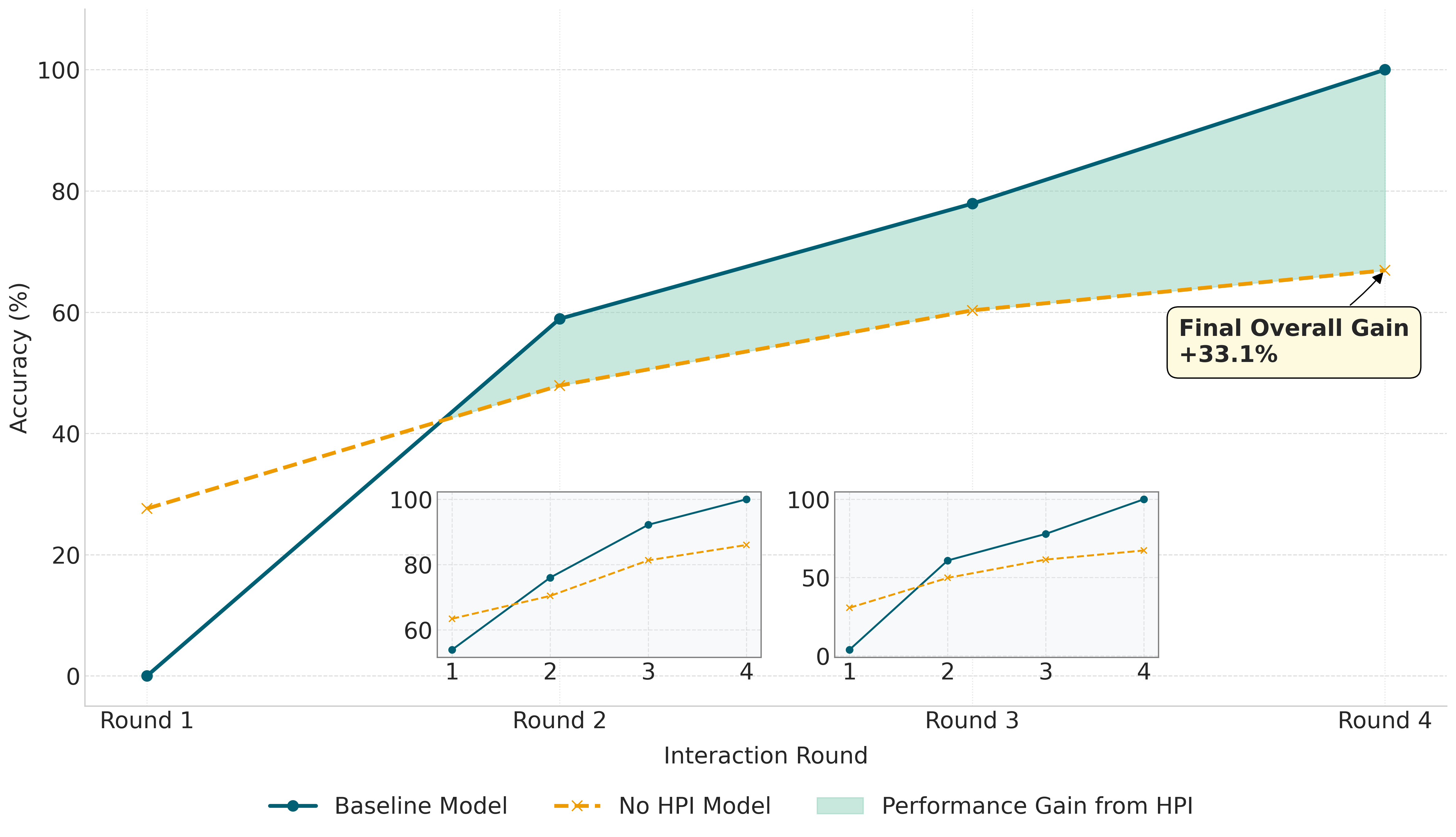} 
    \caption{Structured HPI Unlocks Diagnostic Capabilities on Challenging Cases. The ablation study shows that including a structured History of Present Illness (HPI) enables the baseline model to significantly outperform the model without HPI. The performance gap widens with each interaction round, culminating in a +33.1\% final accuracy gain.}
    \label{fig:figure6}
\end{figure}

The results, visualized in Figure~\ref{fig:figure6}, present a stark contrast. Our Model demonstrates a remarkable self-correction capability, with its overall accuracy soaring from 0\% in the first round to a perfect 100\% by the final round. In contrast, the No HPI Model, deprived of structured information, loses its ability to solve complex problems almost entirely; its learning curve is exceptionally flat, ultimately stagnating due to an inability to learn effectively. This results in a final performance gap of +33.1\%, a figure that powerfully quantifies the core value created by the RecipientAgent.This experiment proves that the RecipientAgent is not a mere incremental improvement but a critical, foundational component of our architecture. 

\subsection{Effectiveness of Intelligent Guidance Mechanisms}
To evaluate the value of our two intelligent guidance mechanisms---the Inquiry Guidance Mechanism and the Classification Guidance Mechanism---we conducted a comprehensive ablation study. The results, presented in Table~\ref{tab:guidance_main_comparison}, unequivocally demonstrate the indispensable synergistic value of these two mechanisms. Our model which integrates both strategies, achieved a perfect 100\% accuracy, with its Overall Accuracy Gain reaching a remarkable +100\% percentage points. In contrast, the No Guidance model improved by only +18.6\%, indicating that without effective guidance, the model cannot learn efficiently from interaction, even when provided with a high-quality HPI.

This performance gap stems from a stark difference in inquiry efficiency in Table~\ref{tab:guidance_efficiency}. Models equipped with the Inquiry Guidance Mechanism continuously improved with each interaction, achieving 100\% inquiry efficiency. Conversely, models lacking this guidance stagnated after the second round, with their efficiency dropping to 67\%. These findings prove that the Inquiry Guidance Mechanism is crucial for ensuring corrective momentum and efficiency, while the Classification Guidance Mechanism provides logical rigor. 

\begin{table}[htbp]
\centering
\renewcommand{\arraystretch}{1.2}
\setlength{\tabcolsep}{5pt}
\small
\begin{tabular}{l rrr r}
    \toprule
    \textbf{Model} & \textbf{Overall Acc.} & \textbf{Prim. Acc.} & \textbf{Sec. Acc.} & \textbf{Gain} \\
    \midrule
    \textbf{IG + CG} & \textbf{100.0} & \textbf{100.0} & \textbf{100.0} & \textbf{+100.0} \\
    IG & 62.0 & 81.4 & 62.8 & +31.8 \\
    CG & 57.8 & 80.6 & 59.3 & +26.4 \\
    NG & 45.3 & 73.3 & 48.1 & +18.6 \\
    \bottomrule
\end{tabular}
\caption{Triage Accuracy Comparison Across Guidance Strategy Configurations. IG+CG represents our model (Inquiry Guidance + Classification Guidance), IG: Inquiry Guidance only, CG: Classification Guidance only, NG: No Guidance.}
\label{tab:guidance_main_comparison}
\end{table}
\vspace{1em} 

\begin{table}[htbp]
\renewcommand{\arraystretch}{1.2}
\begin{tabular}{l c}
    \toprule
    \textbf{Model Configuration} & \textbf{Effective Inquiry Rounds (\%)} \\
    \midrule
    \textbf{IG + CG} & \textbf{100\%} (3 of 3) \\
    IG & 100\% (3 of 3) \\
    CG & 67\% (2 of 3) \\
    NG & 67\% (2 of 3) \\ 
    \bottomrule
\end{tabular}
\caption{Inquiry Efficiency of Different Guidance Strategies.}
\label{tab:guidance_efficiency}
\end{table} 

\bibliography{aaai2026}

\end{document}